\documentclass[10pt,conference]{IEEEtran}
\usepackage{graphicx,
	psfrag,
	epsfig,
	epstopdf,
	amsthm,
	amssymb,
	url,
	subcaption,
	balance,
	enumerate,
	color,
	setspace,
	tikz,
	pgfplots
}
\usepackage{amsmath}
\usepackage[nospace,noadjust]{cite}
\usepackage{pbox}
\usepackage{multirow}
\usepackage{adjustbox}
\usepackage{array}
\usepackage{booktabs}
\usepackage{float}
\usepackage{breqn}
\usepackage{amsmath}
\usetikzlibrary{shapes.geometric}

\newtheorem{example}{Example}
\newtheorem{definition}{Definition}
\newtheorem{problem}{Problem}
\newtheorem{lemma}{Lemma}
\newtheorem{observation}{Observation}
\newtheorem{theorem}{Theorem}
\newtheorem{corollary}{Corollary}
\newtheorem{proposition}{Proposition}
\newtheorem{argument}{Argument}
\newtheorem{result}{Result}
\newtheorem{note}{Note}
\newtheorem{claim}{Claim}
\newtheorem{property}{Property}

\newcommand{\cref}[1]{Constraint~\ref{#1}}

\newcommand{\ignore}[1]{}
\usepackage[paperwidth=8.5in, paperheight=11in,top=1.1in, bottom=0.99in, left=0.61in, right=0.61in]{geometry}

\usepackage{algpseudocode}
\usepackage{algorithm}
\usepackage{comment}

\begin{document}


\title{Implicit Sensing in Traffic Optimization: Advanced Deep Reinforcement Learning Techniques}

	\author{
	\IEEEauthorblockN{ Emanuel Figetakis\IEEEauthorrefmark{1}, Yahuza Bello \IEEEauthorrefmark{1}, Ahmed Refaey \IEEEauthorrefmark{1}\IEEEauthorrefmark{2}, Lei Lei \IEEEauthorrefmark{1}, and Medhat Moussa \IEEEauthorrefmark{1}}\\

	\IEEEauthorblockA{\IEEEauthorrefmark{1} University of Guelph, Guelph, Ontario, Canada.}\\
	\IEEEauthorblockA{\IEEEauthorrefmark{2} Western University, London, Ontario, Canada.}}

\maketitle
\begin{abstract}
A sudden roadblock on highways due to many reasons such as road maintenance, accidents, and car repair is a common situation we encounter almost daily. Autonomous Vehicles (AVs) equipped with sensors that can acquire vehicle dynamics such as speed, acceleration, and location can make intelligent decisions to change lanes before reaching a roadblock. A number of literature studies have examined car-following models and lane-changing models. However, only a few studies proposed an integrated car-following and lane-changing model, which has the potential to model practical driving maneuvers. Hence, in this paper, we present an integrated car-following and lane-changing decision-control system based on Deep Reinforcement Learning (DRL) to address this issue. Specifically, we consider a scenario where sudden construction work will be carried out along a highway. We model the scenario as a Markov Decision Process (MDP) and employ the well-known DQN algorithm to train the RL agent to make the appropriate decision accordingly (i.e., either stay in the same lane or change lanes). To overcome the delay and computational requirement of DRL algorithms, we adopt an MEC-assisted architecture where the RL agents are trained on MEC servers. We utilize the highly reputable SUMO simulator and OPENAI GYM to evaluate the performance of the proposed model under two policies; $\epsilon$-greedy policy and Boltzmann policy. The results unequivocally demonstrate that the DQN agent trained using the $\epsilon$-greedy policy significantly outperforms the one trained with the Boltzmann policy.

\end{abstract}

\begin{IEEEkeywords}
Lane-changing, car-following, MDPs, DQN, Autonomous vehicles, Intelligent transportation systems
\end{IEEEkeywords}

\section{Introduction}
In the past two decades, researchers have been working on innovative ways to make significant improvements in security and comfort in Intelligent Transportation Systems (ITS) \cite{8771378}. Safety is one of the major design goals for autonomous driving as it allows autonomous vehicles to navigate roads independently with fewer human interventions. This will eventually lead to fewer accidents compared to human drivers who are mostly impaired due to many reasons such as sickness and fatigue after a long drive to name a few. A multitude of onboard sensors is used in most current autonomous driving systems to gather relevant data \cite{9617150} \cite{trackingsensors}. Sharing these relevant data among multiple autonomous cars will be beneficial to have an efficient ITS. The advent of advanced communication technologies such as Vehicle-to-Vehicle (V2V), Vehicle-to-Infrastructure (V2I), and Vehicle-to-everything (V2X) allows for highly efficient wireless communication within the ITS domain \cite{ouaissa2022secure}. This is very important for autonomous vehicles to communicate with one another, central servers, and off-board Road Side Units (RSUs). 

An essential aspect of autonomous vehicle operation in complex driving scenarios is having a decision-making control module that is accurate and that can be executed almost instantly. This module is responsible for sending instructions to the vehicle's action execution module to perform numerous actions like following cars, avoiding obstacles, changing lanes, and overtaking \cite{rosenzweig2015review}. Driving maneuvers such as following a car in front and changing lanes are the two most important driving situations that occur most frequently. Consequently, multiple research works have emerged with various car-following models and lane-changing models \cite{9439506,9057371,9468359,9810922,9837330,9843863}. However, most of these research works studied and proposed car-following models and lane-changing models individually. The effects of lane-changing behavior on vehicles in the opposite lane have been pointed out by many scholars. To this end, the importance of joint research on vehicle driving systems that considers both lane-changing behavior and car-following behavior cannot be overstated.

Recently, Machine Learning (ML) based car-following models have yielded outstanding performance as reported in the literature. Specifically, Reinforcement Learning (RL) and Deep Reinforcement Learning (DRL) stand out among other ML approaches adopted \cite{9439506,9057371,9468359}. Similarly, lane-changing models are seeing similar research trends in the literature (i.e., researchers are adopting RL and DRL such as Deep Deterministic Policy Gradient (DDPG) and Deep Q-Network (DQN) algorithms to solve the lane-changing maneuver problem) \cite{9810922,9837330,9843863}. 
However, only a few research works have been reported in the literature \cite{peng2022integrated}. 

For any ITS to be effective, vehicle-to-everything (V2X) applications must be deployed to enable vehicles to exchange information with nearby vehicles and infrastructure to coordinate maneuvers. It is very common to execute several intensive computational operations in a very short period of time in the ITS domain to have safe and effective coordination among vehicles. This necessitates the deployment of servers with high computational resources along the road. As part of the V2X infrastructure deployment, operators install RSUs to reduce communication delays between vehicles and central servers, which improves the coverage range.  

In response to concurrent delay and computational requirements, the European Telecommunications Standards Institute (ETSI) introduced the multi-access edge computing (MEC) concept \cite{isg2018multi}. With MEC, computational resources are moved closer to the vehicles. MEC-assisted ITS applications are being heavily investigated in the literature. 

The contributions of this paper are summarized as follows: 
\begin{itemize}
\item Develop a cohesive decision control framework for car-following and lane-changing operations, utilizing DRL techniques. This is specifically tailored for scenarios involving abrupt highway construction work.
\item Formulate the given scenario as an MDP and employ the Deep Q-Network (DQN) algorithm during the experimentation phase to train an RL agent in making optimal decisions.
\item Integrate a MEC-assisted architecture to address latency and computational demands associated with DRL algorithms.
\item Evaluate and contrast two distinct decision-making policies, namely Boltzmann and Epsilon Greedy, to ascertain the superior approach in enhancing traffic flow efficiency within the simulation environment.

\end{itemize}


The structure of the rest of the paper is as follows: Section II presents the background knowledge and the relevant related works in the literature. Section III presents the proposed system model and its description. Section IV presents the description of the implementation of the environment and model used for training the agent. The section concludes with results and a discussion of the performance analysis of the proposed model.
Section V gives conclusion remarks and highlights the intended future work.


\begin{figure*}[ht!]
   \includegraphics[width=\textwidth,height=68mm]{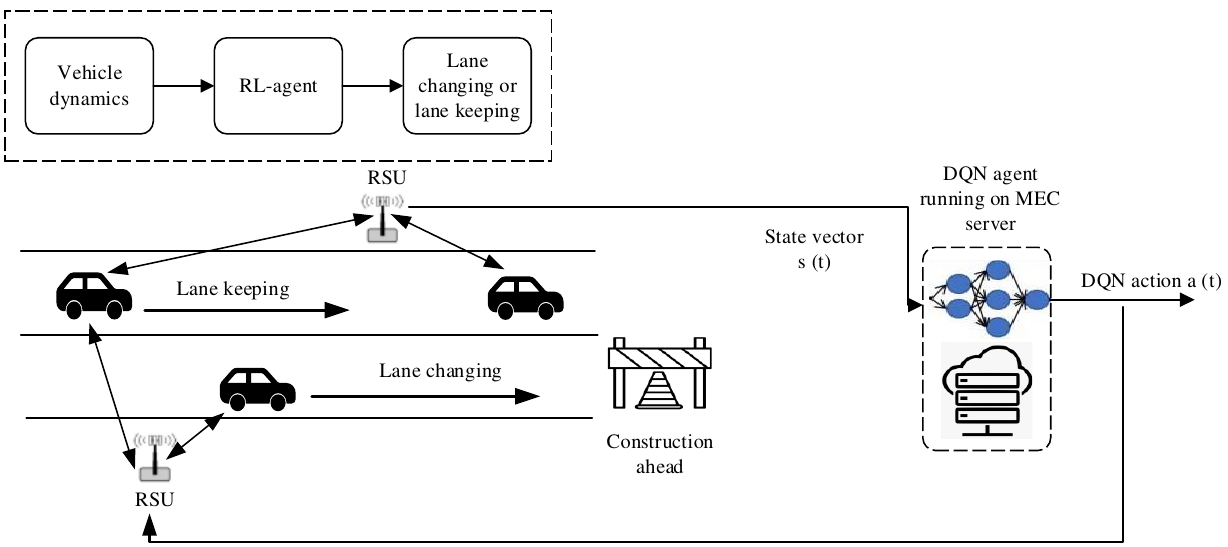}
    \caption{Proposed integrated car-following and lane-changing model running on MEC server}
    \label{fig:PM}
\end{figure*}

\section{Background and Related Works}

The purpose of this section is to discuss the necessary background knowledge required and the related works from the literature. The first part discusses the concept of RL and DRL in general. The second part dive into the car-following models, and lane-changing models.

\subsection{RL and DRL}

Generally, RL describes an autonomous agent learning by interacting with its surroundings to improve its performance. A reward function $R$ determines whether an agent performs well or not in a typical RL environment. A decision is made for every state experienced and an occasional reward, $R$, is received as a result based on the usefulness of that decision by the agent. In order to maximize cumulative rewards over a specified period of time, the agent must maximize each reward received during the specified period, which is the ultimate goal of the agent. As the agent learns more about the expected utility of different state-action pairs (i.e. discounted sum of future rewards), it can gradually increase its long-term reward.

When formalizing sequential decision problems with an RL agent, Markov decision processes (MDPs) seem to be the standard that is widely adopted \cite{wang2022deep}. A typical MDP is characterized as a tuple $<S, A, P, R, \gamma>$ where $S$ is the set of state space, $A$ is the set of action space, $P$ is the probability transition function, $R$ is the reward function, and $\gamma$ is a discount factor used to adjust immediate and future rewards.


In short, when an agent in a state $s \in S$ chooses an action $a \in A$, the agent will transition to the next state $s' \in S$ according to the transition probability $P \in [0,1]$ and gain a reward $R(s,a)$. Note that the end goal of an agent is to find the optimal policy $\pi^*$ (which is a mapping of the environment states to actions) that will result in the maximum expected sum of discounted rewards \cite{wang2022deep}. The agent uses this policy to determine how to behave at every step of the way. A policy can either be deterministic or stochastic in nature. A deterministic policy typically returns the action to be taken in each state whereas as a stochastic policy returns a probability distribution over the actions (usually donated as $\pi(a|s)$). Based on a given policy, a value-function or a Q-function can be defined to measure the expected rewards starting from any given state $s \in S$ or any state-action pair $(s,a)$. For more details on the different RL methods for finding the optimal value-function, Q-function, and policy, refer to \cite{9351818}.

Despite the fact that learning in RL is a convergent process, it takes a long time to come up with the best policy because it requires a lot of exploration and knowledge, so it is not suitable and inapplicable for large-scale networks. This limits the wide adoption of RL in many applications. The advancement of deep learning paves the way for a new RL approach called DRL, which solves some of the limitations of the traditional RL. By utilizing Deep Neural Networks (DNNs) as a learning tool, DRL is able to increase the learning speed and performance of RL algorithms. Consequently, DRL is being adopted in many applications such as ITS, healthcare, robotics, trading and finance, news recommendation platforms, and networking. Several algorithms such as DQN, DDPG, and Double Deep Q Network (DDQN) emerged as a result of the advancement in the DRL paradigm \cite{9904958}.

\subsection{Car-following models and lane-changing models}
Several studies emerged in the domain of ITS tackling the problem of car-following and lane-changing decisions. A car-following framework for Autonomous Vehicles (AVs) is developed in \cite{9439506} using a navigation decision module and an automatic object detection module. The authors explore Q-learning and DQN algorithms against the proposed car-following framework. In \cite{9057371}, the authors explored the characteristics of demand for AVs in mixed traffic flow and proposed a car-following model with a warning module. An automated entropy adjustment algorithm using Tsallis actor-critic (ATAC) is established in \cite{9468359} to propose a car-following model. The authors explore Twin Delayed DDPG (TD3), Soft
Actor-critic (SAC) and DDPG algorithms to test the performance of the proposed car-following model.

Similarly, numerous research works have emerged in the domain of lane-changing maneuvers. The authors in \cite{9810922} proposed a lane-changing decision module for AVs using a game theoretical approach. The authors demonstrated that during the course of the lane-changing decision, the objective quantification of lane-changing intention must be used in order to determine the lane-changing intentions of AVs. The authors in \cite{9837330} proposed an improved method for planning the trajectory of changing lanes for AVs on closed highways based on segmented data. For predicting target vehicles' driving intentions, a Bayesian Network-based model using Long Short-Term Memory and the maximum entropy Inverse Reinforcement Learning (IRL) algorithm is used. 

Some research works that explore the integration of car-following and lane-changing models for better decision options are starting to trend in the literature. For example, the authors in \cite{peng2022integrated} proposed a two-layered framework that utilizes DRL algorithms to handle large-scale mixed-state spaces and produces lane-changing and car-following decisions as a composite output. A Duelling Double Deep Q-Network (D3QN) algorithm is used to distinguish the value of selected lane-changing actions and the potential value of the environment in the upper layer model. A DDPG algorithm is used in the second layer of the model to control vehicle speed continuously based on car-following decisions. This paper differs from the work in \cite{peng2022integrated} in the sense that we use a single RL algorithm (i.e., DQN algorithm) to make decisions on car-following or lane-changing actions. Moreover, the work in \cite{peng2022integrated} did not adopt a MEC-assisted architecture, rather assuming the vehicle is equipped with the required processing resources to run the RL agents.



\section{System Model}

The proposed integrated car-following and lane-changing model is presented in Figure~\ref{fig:PM}. We consider a scenario where a sudden construction or roadblock is up ahead on a highway. To model the scenario properly, we assume the autonomous vehicle is equipped with multiple sensors and appropriate wireless communication systems such as V2V or V2I systems, which can be depicted in Figure~\ref{fig:PM} as the vehicle dynamic module. This means that the target vehicle can be able to gather information such as its own speed, acceleration, position, distance to the roadblock, and the same information about neighboring vehicles. We also assume to have off-board Road Side Units (RSUs) that the vehicles can exchange data. The vehicles' data/information is passed through a DQN algorithm to train the agent to make appropriate decisions regarding lane-changing or car-following. The DQN algorithm is implemented on a central MEC edge server. This central edge server handles the computational calculations of the MDP problem and then communicates the decision to the vehicles accordingly through V2I technology using the nearby RSU.

Mathematically, we model the scenario as an MDP as follows:

\begin{itemize}
    \item \textbf{State space definition:} We model each state $s \in S$ as tuple $<d_c, d_f, v_{EH}, x_p, y_p >$ where $d_c$ represents the distance of the target car from the roadblock, $d_f$ represents the distance of the target car to the next car in its path, $v_{EH}$ represents the speed of the target car, $x_p$ and $y_p$ represents the location of the target car along the x-axis and y-axis respectively. 
    \item \textbf{Action space definition:} The action space $a \in A$ is given as $< M, C>$ where $M$ represents the decision of merging to the other lane (i.e., lane-changing decision) and avoiding reaching the roadblock and $C$ represents the decision of staying on the same lane (i.e., a car-following decision for the cars on the right lane or cars that are far away from the roadblock). Note that this action space is represented by 12 discrete action spaces, which correlate to the positions in the lane with the roadblock where lane-changing action can be initiated. This will be explained in more detail in section IV.
    \item \textbf{Reward definition:} The reward definition is given as a tuple $<R_M^+, R_M^-, -R_M^+>$ where $R_M^+$ represents a positive reward when the target car successfully changes lane and merge with average speed before reaching the roadblock, $R_M^-$ represents a negative reward when the target car fails to change lane after reaching the roadblock (i.e., the end of the lane in this case) and $-R_M^+$  represents negative positive reward when the target car change lane and merge with minimum speed before reaching the roadblock. The last reward will deal with the situation where the target car eventually merges but at the end of the lane (which will cause high traffic jams).

\end{itemize}

To solve the above modeled MDP problem, we utilize the well-known DQN algorithm because it achieves good performance in related decision control problems \cite{9439506}. We employ two policies; $\epsilon$-greedy policy and Boltzmann policy as explained in the DQN algorithm's pseudocode in algorithm \ref{alg:DQN}. The vehicles gather information about the state space as defined in the state space definition and send it to the nearest off-board RSUs. The off-board RSUs forward the state space to the central MEC edge server, which uses the DQN algorithm to determine the correct actions to execute. The current action is then sent to the corresponding vehicle via the nearby RSU. It is worth mentioning that we adopt the two policies (i.e., $\epsilon$-greedy and Boltzmann) for the sake of performance evaluation as shown in algorithm \ref{alg:DQN}.

\begin{algorithm}
\caption{DQN algorithm on the MEC server}\label{alg:DQN}
\begin{algorithmic}[1]
\Require Initial \textbf{w} set randomly, Initial replay buffer $D$ set to $N$, Initial policy $\pi(s)$ as either $\epsilon$-greedy policy or Boltzmann policy, Initial $S_0$

\For{$t = 0$ to $T$}  
    \State Execute action $A_t = \pi(S_t)$ according to either $\epsilon$-greedy policy or Boltzmann policy
    \State Observe state $S_{t+1}$ and reward $R_{t+1}$
    \State Store the transition parameters $S_t, A_t, R_{t+1}, S{t+1})$ in $D$
    \State Sample minibatch transitions randomly from $D$ as $(S_j, A_j, R_{j+1}, S{j+1})$
    \For{$j$ in minibatch}
      \State Set $y_j=\biggl\{\begin{array}{cl}
      R_{j+1},  S_{j+1}\!\\
      R_{j+1}+\gamma max_{a^'}Q(S_{j+1},a^';w) \end{array}$

      \State Update \textbf{w} using stochastic descent to minimize $(y_j-Q(S_j,Aj;\textbf{w}))^2$
    \EndFor
    \State Improve policy $\pi(s)$ as either $\epsilon$-greedy or Boltzmann with the new \textbf{w}
    \EndFor
\end{algorithmic}
\end{algorithm}

\section{Implementation and Experiment}


The implementation of the system model is a multi-layer problem with the end goal for the extermination to include two RL agents with different policies to take action in the system model. The following tasks are defined in order to achieve the goal of our experimentation:
\begin{itemize}
    \item Model the MDP from the System Model into a Python OpenAi GYM environment
    \item Create the simulation space in Simulation of Urban Mobility(SUMO)
    \item Create the model and add the agent and policy to the Model
    \item Interface the OpenAI GYM environment with the simulation space in SUMO
    \item Interface the Agent with the environment
\end{itemize}

\subsection{GYM Environment}
The GYM environment was modeled after our MDP formulation. The GYM environment allows us to define the parameters of our problem such as state space, action space, observation space, reward function, and simulation length. The library is also not limited to these parameters and can include more complex functions and parameters however, since the GYM environment is being interfaced with SUMO there are fewer parameters needed. The definition of our parameters is as follows, for the state space it is determined by the observation space which is the range that is determined by each lane's average speed. Our action space is a discrete 12-action space that correlates to positions in the lane of construction where a lane change will be initiated. 



\begin{figure}[h]
   \includegraphics[width=.4\textwidth,height=43mm]{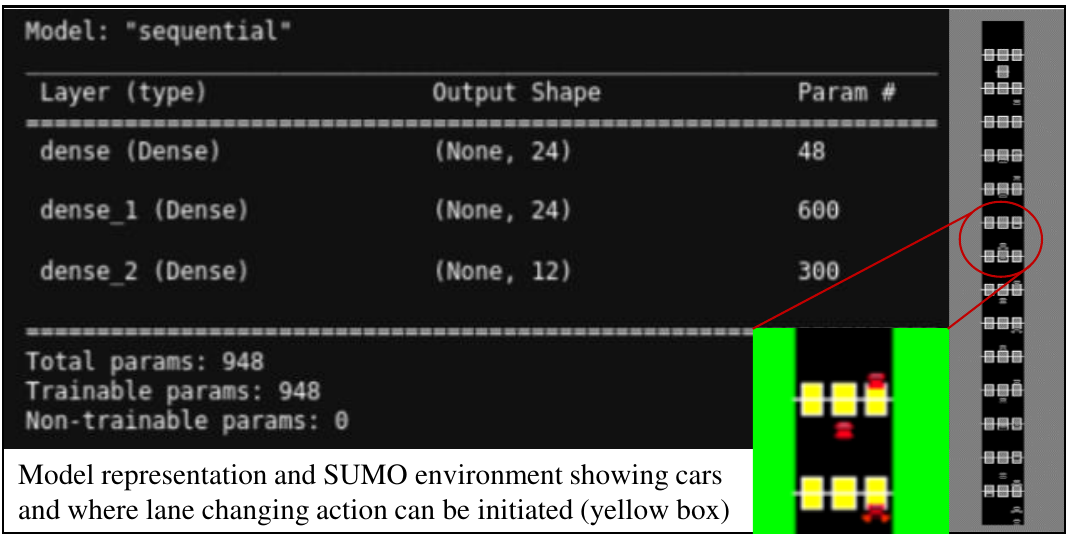}
    \caption{Sumo environment and model representation}
    \label{fig:SUMOEnv}
\end{figure}

\subsection{SUMO Environment}
The SUMO environment is a 300-unit, three-lane highway with a blocked lane at 280 units, causing vehicles to slow down and make lane-changing decisions. A vehicle can change lanes every 20 units in the lane leading up to the roadblock. The RL agent must determine when to start changing lanes to avoid a negative reward if the average speed of the lanes falls below a certain threshold. The work features seamless integration between the two libraries, and more scenarios will be added in future work. Figure \ref{fig:SUMOEnv} depicts the environment with yellow boxes representing positions for lane changes, including one at the construction site and the topmost detectors used to gather information about average speed.

\subsection{Model}

A shallow network with 948 parameters was used to support a DQN model (depicted in Figure \ref{fig:SUMOEnv}) for the GYM environment. It consisted of an input layer based on possible states, followed by a 24-dense layer with ReLu activation, another 24-dense layers, and a model for 12 discrete actions. Testing showed that the program completed one step per second, although this was due to the SUMO simulation rather than the network. The DQN implemented two policies - the Boltzmann Q-Policy and the Greedy Q-Policy - with all other hyperparameters kept consistent. In future works, an actor-critic network could be used to evaluate actions as they occur, which would benefit multi-agent testing.





\subsection{Interfacing Agent, GYM, and SUMO}

At this point in the implementation all modules have been created so comes the time for all modules to work together to create the simulation. The agent that utilizes Tensorflow and Keras interface works with OpenAI GYM without any problems, however, the challenges began when trying to implement SUMO with the agent and the environment. There was no standard way found to implement the two together; the only use factor that was found during research was that SUMO had a Python library integration tool called TraCI. From this Python library the SUMO application can be called from Python, this is what was needed for the implementation. A function was created that took a number between 1 and 12, which corresponds to the point at which the lane change was supposed to take place, and then launched the SUMO simulation with those parameters. 


\section{Simulation Results and Analysis}

This section presents the performance evaluation of the proposed model. For each model, the system was trained for 400 episodes which correlate to 20,000 steps. Several different metrics were recorded and a small noise factor was included in each simulation to simulate randomness. Along with this randomness, the vehicles starting positions were not always the same. With all of this included each policy was able to perform and find optimal solutions.

\subsection{DQN Boltzmann Q-Policy and Epsilon Greedy Q-Policy}

The DQN model with the Boltzmann Q-Policy is intended to take a more exploratory approach. It tries different actions and learns random actions based on a scale of the Q function. The Greedy Q-Policy will quickly find which actions correlate to the highest rewards and based on the state take different actions. 

\begin{figure}[htb]
    \centering
    \includegraphics[width=.4\textwidth]{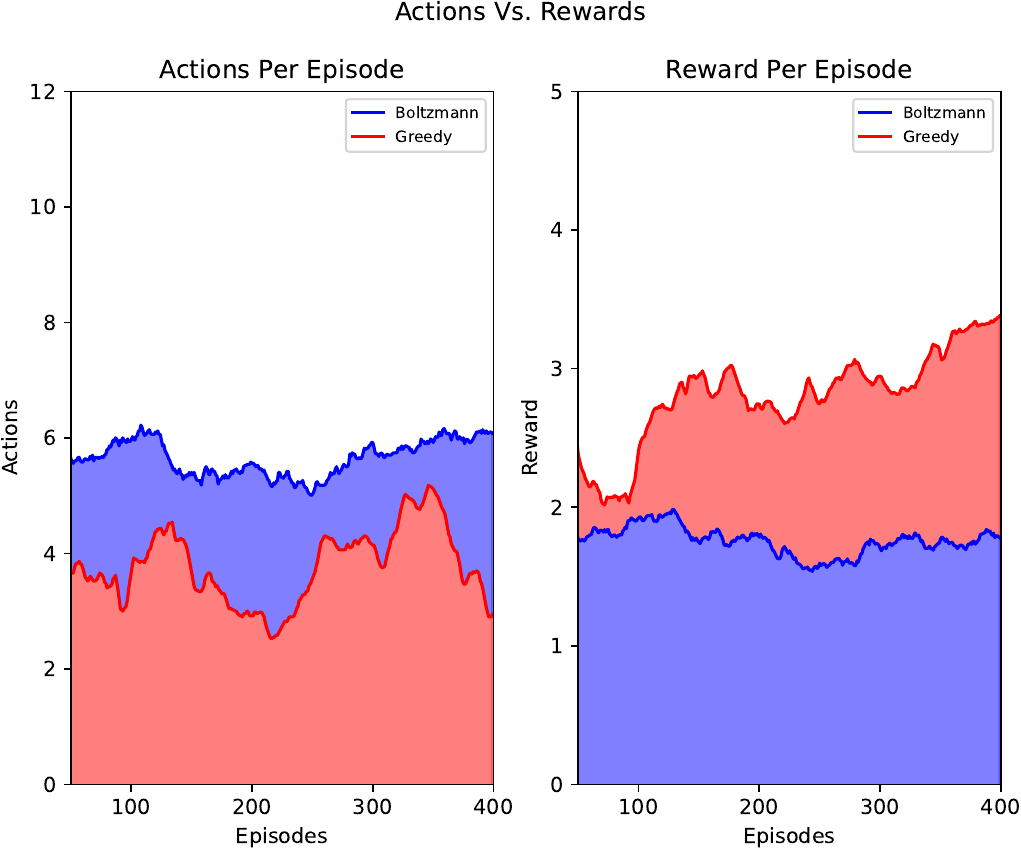}
    \caption{Action and Reward Data}
    \label{fig:Reward}
\end{figure}

Figure \ref{fig:Reward} shows that for this environment the Greedy policy is actually taking more diverse actions whereas Boltzmann is taking the more consistent actions. However, the rewards for the Greedy Policy are greater than the Boltzmann, this is because they are working in a dynamic environment where the optimal actions change per step. The Greedy policy is able to learn this and adapt whereas Boltzmann is stuck evaluating random actions for every scenario on its scale. The problem that the Boltzmann policy is facing is analogous to an optimization problem where gradient descent is getting stuck in local minima.

\begin{figure}[htb]
    \centering
    \includegraphics[width=.4\textwidth,height=58mm]{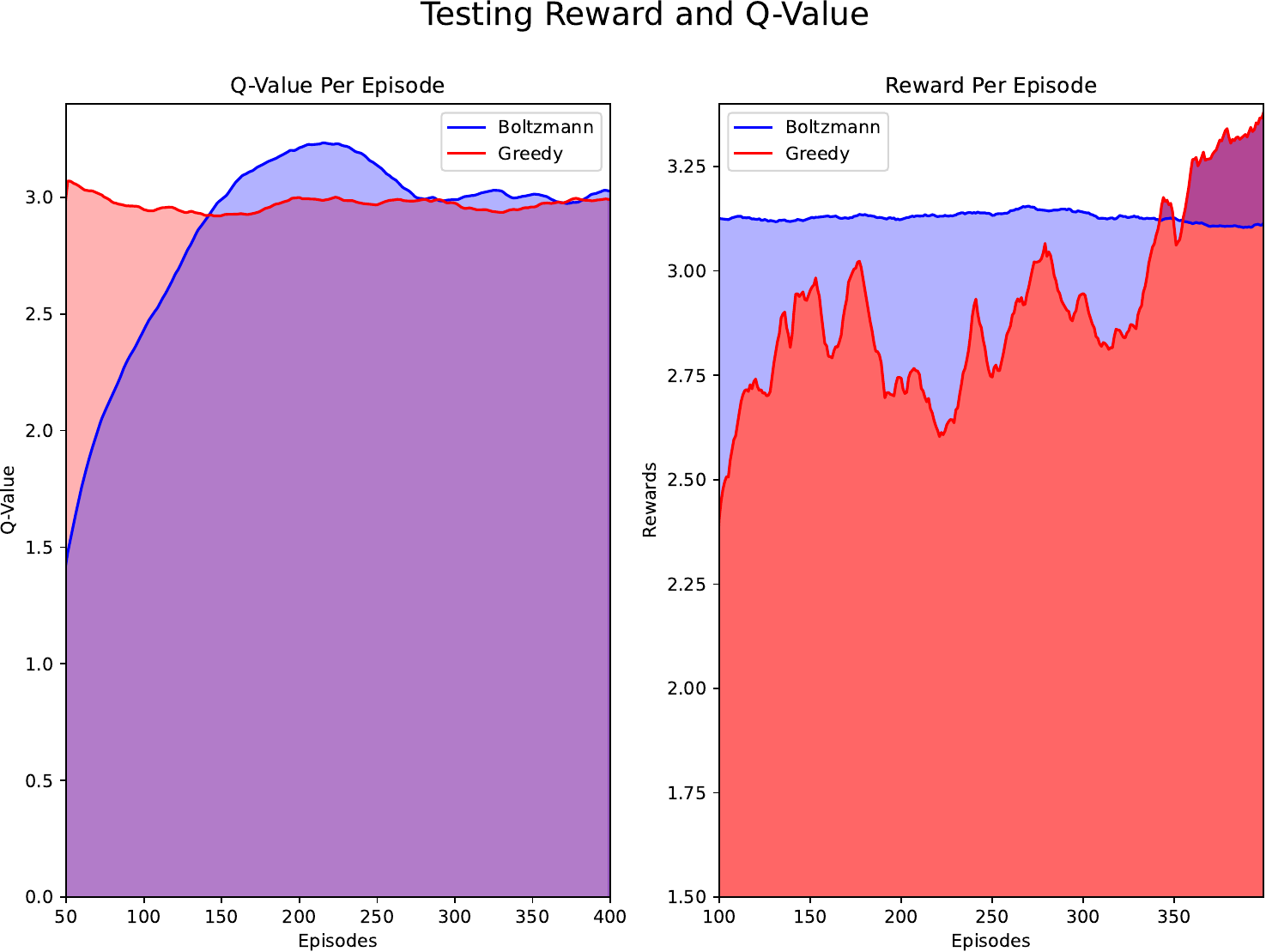}
    \caption{(Q-Value Function at each episode and Rewards of Both Models}
    \label{fig:Q-val}
\end{figure}

Figure \ref{fig:Q-val}(a) helps us understand why the Boltzmann policy is operating poorly compared to the Greedy policy. The Q-Values of Boltzmann are converging with the Greedy policy however, figure \ref{fig:Reward} shows that this is not the case otherwise the rewards would be similar. The Boltzmann Policy is evaluating its actions Q-Values as good but since the environment is dynamic and is changing after every step this is not the case. This is due to the limited memory of the algorithm and is shown by Figures \ref{fig:Q-val}(a) and \ref{fig:Reward}.

After both models were trained they were saved and used to play in the environment. Before running both models, based on the results that were received during testing, we predicted that the Greedy policy would outperform the Boltzmann policy. Looking at figure \ref{fig:Q-val}(b), shows that our prediction was correct. The Boltzmann policy takes more consistent actions that result in the same rewards, while the Greedy Q starts taking different actions to increase rewards as the simulation runs. This is because the Boltzmann Q operates on a range of actions that it determines based on Q values, while the Greedy Q takes actions with the intention of increasing rewards and therefore Q Value. Another reason the Greedy Q outperforms the Boltzmann later in the simulation is the discount factor in the equation as it focuses on gathering rewards over a longer period of time. 

\begin{figure}[htb]
    \centering
    \includegraphics[width=.45\textwidth,height = 50mm]{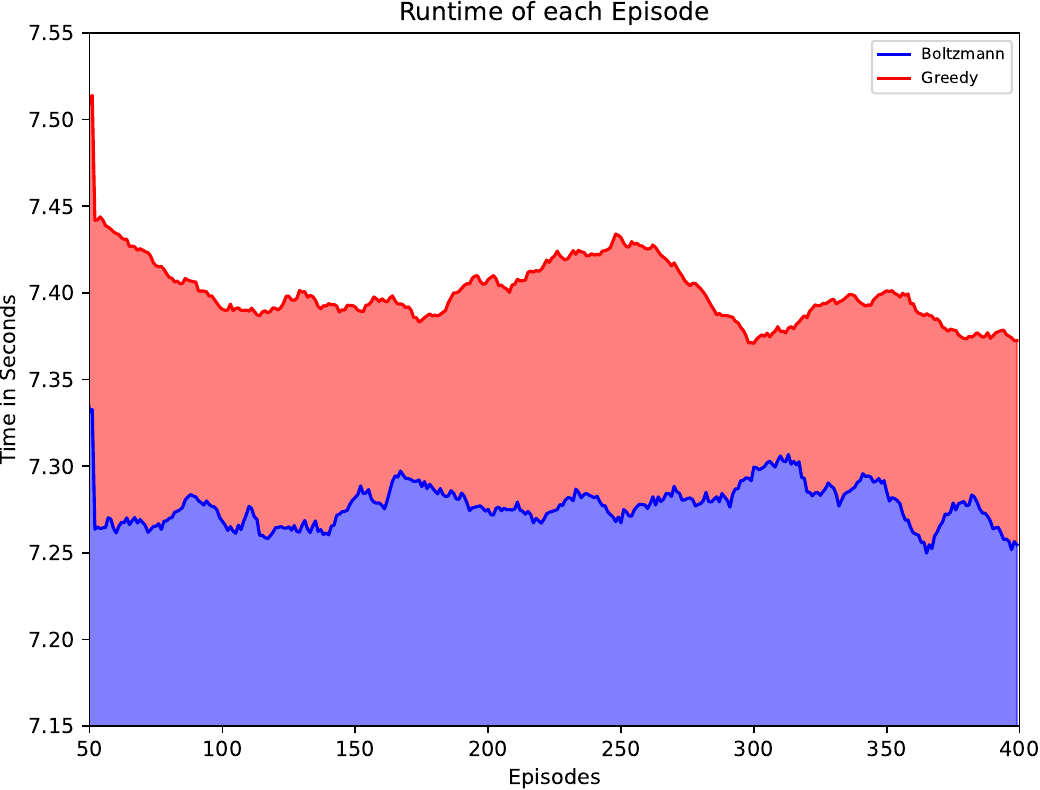}
    \caption{Run time of Training both DQN Agents}
    \label{fig:Runtime}
\end{figure}

Figure \ref{fig:Runtime} shows the program created for this implementation has a reasonable run time. It also shows that Greedy policy is taking a little more time and this can be because the algorithm includes memory for its previous steps before making a new action whereas Boltzmann is making different actions every step. 

\section{Conclusion and Future Work}

In this paper, the task of modeling traffic and optimal merge patterns was accomplished through the use of Deep Reinforcement Learning. This task presented many challenges that required creativity when creating the training and testing programs. The problem had to be modeled into a game environment, and then the RL algorithm had to be trained and tested to gather results. It features a unique and novel implementation with Python, OpenAI GYM, and SUMO. With the use of this novel program, two different RL policies were able to be tested. From the experiment, it can be determined for the application of merge patterns the Epsilon Greedy Q-Policy performs better than the exploratory Boltzmann Q-Policy. Also, due to the dynamic nature of the environment that was created, the conclusion is that Greedy Q performs better in a random environment. 
Since the framework for the problem has been created this presents opportunities for different algorithms and policies to be tested. The framework can be scaled to almost any RL or ML framework. We hope to further contribute by evaluating different algorithms to test performance against one another, as well as perform tweaks to the environment to add several different scenarios.

\bibliographystyle{IEEEtran}
\bibliography{references}
\end{document}